\documentclass[runningheads]{llncs}
\usepackage[T1]{fontenc}
\usepackage{threeparttable}  
\usepackage[misc,geometry]{ifsym}
\usepackage[linesnumbered,ruled]{algorithm2e}
\usepackage{amsmath}
\usepackage{caption}
\usepackage{subcaption}

\usepackage{amsfonts}
\usepackage{graphicx}
\begin{document}
\title{SPGNet: Spatial Projection Guided 3D Human Pose Estimation in Low Dimensional Space}
%
%
\author{Zihan Wang\inst{1} \and
Ruimin Chen \inst{1} \and
Mengxuan Liu\inst{1} \and
Guanfang Dong \inst{1} \and
Anup Basu \inst{1}
}

\authorrunning{Z. Wang et al.}
\titlerunning{Spatial Projection Guided 3D Human Pose Estimation}

%
\institute{University of Alberta, Edmonton, Canada \\  
\email{\{zihan6, ruimin6, mliu2\}@ualberta.ca}, \\}
\maketitle              
\begin{abstract}
We propose a method SPGNet for 3D human pose estimation that mixes multi-dimensional re-projection into supervised learning. In this method, the 2D-to-3D-lifting network predicts the global position and coordinates of the 3D human pose. Then, we re-project the estimated 3D pose back to the 2D key points along with spatial adjustments. The loss functions compare the estimated 3D pose with the 3D pose ground truth, and re-projected 2D pose with the input 2D pose. In addition, we propose a kinematic constraint to restrict the predicted target with constant human bone length. Based on the estimation results for the dataset Human3.6M, our approach outperforms many state-of-the-art methods both qualitatively and quantitatively.
 
\end{abstract}

\section{Introduction}
3D human shape and posture estimation from a single image or video is a fundamental topic in computer vision. Unfortunately, it is not easy to estimate 3D body shape and posture directly from monocular images without any 3D information. The problem of 3D human shape and posture estimation can be defined as giving images as the input, and generating a 3D skeleton as the output. A typical 3D skeleton consists of 3D points for 17 joints. Mathematically it can be written as a mapping function \(f(M) = x\), where \(M\) is a fixed-sized matrix representing the image input and \(x\) is a matrix representing the 3D joints (size is (17, 3) in our case). Generally, \(f\) is applied to each frame of the video.

Before deep learning was widely used, massive 3D labeled data and 3D parameters with prior knowledge were necessary to deal with this problem \cite{Andriluka_20110_IEEE}. After introducing the deep learning strategy, some approaches extract the 3D human pose based on the input image directly without any intermediate stage. Some of the existing strategies rely on convolutional neural networks to learn visual representations successfully from a very large dataset \cite{Sun_2018_ECCV,Pavlakos_2017_CVPR}. However, recent research has shifted to two-stage approaches. The 2D key points detection algorithm is first used to acquire the 2D poses from images. Then, 2D-to-3D pose lifting is applied as the second stage \cite{Chen_2016_CVPR,Pavlakos_2018,MartinezHRL17}. Existing methods have focused on optimizing the loss functions or neural network structures \cite{Chen_2016_CVPR,Pavlakos_2018,MartinezHRL17,temporal_convolutions_and_semi_supervised}. 

Inspired by the cycle consistency in unsupervised learning, some approaches re-map the predicted 3D poses to 2D poses in a semi-supervised learning framework \cite{temporal_convolutions_and_semi_supervised}. Like following the CycleGAN Loss, the network is designed incorporating two components. One is the mapping from 2D to 3D. Another one is re-mapping 3D to 2D and comparing the re-projected 2D poses with the 2D input. However, the results for unlabeled videos are relatively unremarkable and the training process is relatively slow. In our work, we also extract the intermediate 2D pose. However, we do not intuitively deploy the CycleGAN semi-supervised learning framework, which leads us to avoid the abovementioned drawbacks.

\begin{figure}[!t]
\includegraphics[width=\textwidth]{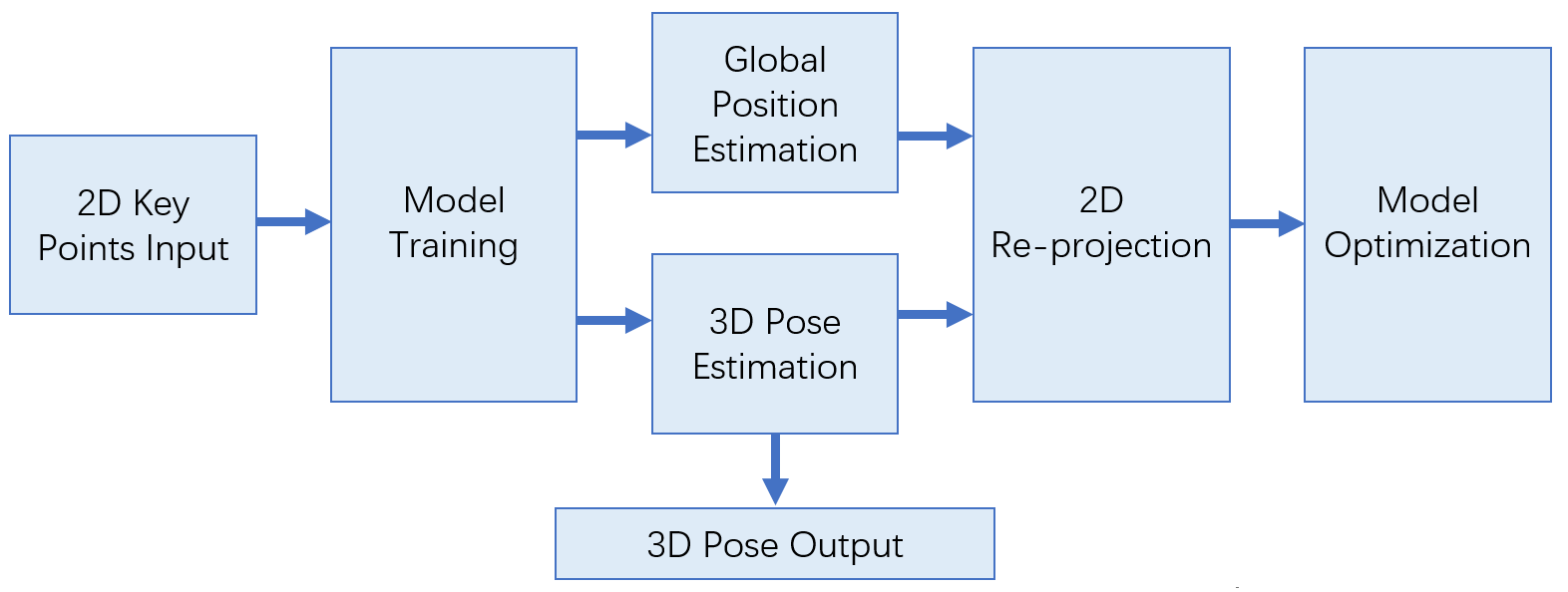}
\caption{General workflow of our SPGNet, take 2D key points as input, and estimate the 3D pose with its global position. The 2D re-projection combines both estimations and loss to optimize the model.} \label{flow_chart}
\end{figure}

We design SPGNet, a novel neural network architecture that combines traditional supervised learning with a spatial projection that re-projects the predicted 3D poses to 2D poses. Instead of traditional loss functions based on the ground truth and 3D pose output, we introduce the 2D MPJPE loss (defined as Equation \ref{eqn:MPJPE}). As shown in Fig. 1, SPGNet computes the 2D projection of 3D poses output with the estimated global position. The 2D MPJPE loss minimizes the re-projected 2D poses with the 2D poses input, which reuses the 2D poses input during the learning process. This approach increases the accuracy of 3D pose prediction. Overall, our main contributions are:
\begin{enumerate}
    \item Introducing an adaptive supervised training framework for 3D human pose estimation under the category of 2D-to-3D lifting approach. 
    \item  Exploiting the 2D pose input efficiently and improving robustness by presenting a re-projection loss, which is based on global pose estimation from 2D poses and the 2D poses themself.
\end{enumerate}
Our model achieves 45.3 millimeters accuracy in Protocol 1 and 35.7 millimeters in Protocol 2 in the Human3.6M dataset, which are \(0.6\%\) and \(1.4 \%\) relative improvements over previous approaches \cite{Xu_2020_CVPR}.

\section{Related Work}

Previous research on 3D Human Pose Estimation can be classified into two main categories. The first category extracts the 3D human pose based on the input image directly without any intermediate stage. Under this category, some recent approaches rely on convolutional neural networks to learn visual representations successfully from very large datasets. The predicted accuracy has increased considerably during the past few years \cite{Sun_2018_ECCV,Pavlakos_2017_CVPR}. The second category consists of two stages, namely 2D key points detection from video and 2D-to-3D pose lifting \cite{Chen_2016_CVPR,Pavlakos_2018,MartinezHRL17}.

\subsection{2D Key Joint detection} Before estimating 3D human pose, many methods require labeled key joints. Inaccurate key joint labels may cause pose prediction to fail. Initially, 2D human pose was estimated based on Deep Neural Networks (DNNs), later Convolutional Neural Networks (CNNs) have shown more advantages in 2D key joint detection \cite{Wei_2016_CVPR,Tompson_2015_CVPR,Newell_2016_CVPR,Yang_2016_CVPR,Toshev_2014_CVPR}. CNN-based models have outstanding performances in extreme test conditions (body occlusion or low-resolution images). A heatmap is generated to show the possibilities of a specific joint shown in an image and the estimate is refined to improve localization. Reusing the hidden layer in the CNN-based model, Tompson et al. proposed a method that uses heatmap, works as a regularizer to find their output, and increases the accuracy \cite{Tompson_2015_CVPR}. A similar method proposed by Yang et al. uses an end-to-end mixture of parts model \cite{Yang_2016_CVPR}. In their method, the probability is calculated by the softmax function and a Max-sum algorithm is applied between pose parts. The pose machine uses multi-stage differentiable iterations to the joint on the heat-map to finally converge to one solution \cite{Wei_2016_CVPR,Newell_2016_CVPR}. The pose machine combines the previous output and the updated prediction for the same input image in each subsequent stage. 

\subsection{Image to 3D pose} Without estimating the 2D human pose, the 3D pose can be constructed directly through an image. This method minimizes the effects of error prediction during the 2D pose estimation, and in general, it can be more robust \cite{Wang_2021_CVPR,Ruiz_2018_CVPR_Workshops,Kanazawa_2018_CVPR}. Additionally, it eliminates the limitation of the unlabeled image used for training \cite{Yang_2018_CVPR}. One of the methods uses Pose Orientation Net (PONet) and generates heat maps: limb confidence maps and 3D orientation maps. With these heat maps, the model uses a fixed-length skeleton to match with the 3D orientation maps and complete the missing limbs in a 3D pose using sub-networks \cite{Wang_2021_CVPR}. For a more specific pose, Ruiz et al. used three loss functions for each angle's rotation, with classification and regression \cite{Ruiz_2018_CVPR_Workshops}. Another approach proposed by Yang et al. uses the Generative Adversarial Networks (GANs) to directly predict the 3D pose using unlabeled images in outdoor scenarios \cite{Yang_2018_CVPR}. Therefore, skipping 2D pose detection provides a solution for unpaired 2D-to-3D data training \cite{Kanazawa_2018_CVPR,Lassner_2017CoRR}.

\subsection{2D-to-3D pose lifting} 3D pose estimation can be lifted from 2D human pose joints \cite{MartinezHRL17,Chen_2016_CVPR,Pavlakos_2018}. During the training, some approaches directly using 2D human pose ground truth as input \cite{Kocabas_2019_CVPR,Drover_2018_ECCVW}. This approach ensures the accuracy of the inputs and distraction is minimized. Moreover, 2D label on the image is easier to obtain. Some datasets, like Human3.6M and HumanEval, contain multiple 2D views, and EpipolarPose with another method estimate 3D pose from different directions at the same frame. The mixture of the multi-views provides consistency loss and recovers the pose via triangulation \cite{Kocabas_2019_CVPR}. The model can be optimized by blending multiple 2D views into the same 3D pose. Depth prediction is important for some 2D-to-3D approaches. The simpler method uses binary ordinal depth relation prediction, while other have explicit depth prediction on each joint \cite{Chen_2016_CVPR,Nie_2017_Monocular3H,Pavlakos_2018}.

\begin{figure}[!htp]
\includegraphics[width=\textwidth]{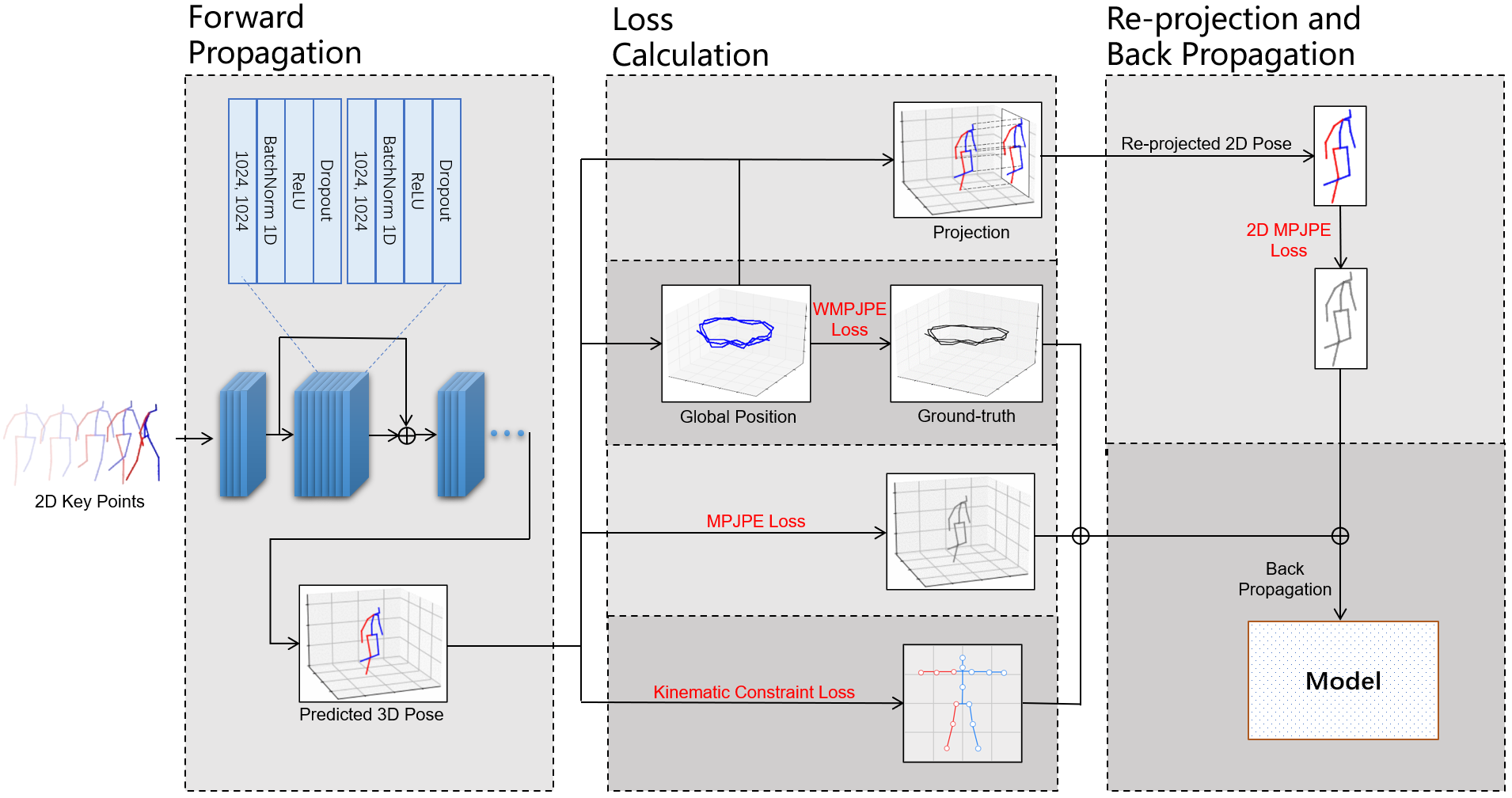}
\caption{Supervised training framework with a 3D pose model with predicted 2D pose sequence input. In addition to the kinematic constraint and MPJPE loss, we concatenate the global position and key points of the 3D poses to perform the projection. Finally, we compare the re-projected 2D pose and original input 2D poses to compute a 2D MPJPE loss and perform backpropagation.} \label{model}
\end{figure}

\section{Proposed Method}

In this section, we briefly explain the details of the proposed SPGNet architecture, as shown in Figure \ref{model}.


\subsection{Problem Formulation}
To better illustrate our method, we first formulate the 3D pose estimation problem as a 2D-to-3D lifting pipeline. Then, we assume the dataset defined as \(\mathcal{D} = \left\{\left(x_i,y_i\right)\right\}^{N}_{i=1}\) consisting of \(N\) data, where each data \(x_i\) is associated with a corresponding label \(y_i\). Here, \(x_i \in \mathbb{R}^{M \times 2}\) represents the one frame input of 2D pose key points. Similarly, \(y_i \in \mathbb{R}^{M \times 3}\) represent estimated 3D pose key points labels, where \(M\) denoted as number of joints for each human pose. In order to take advantage of temporal information between frames, we define the sequence of input for one frame as vector \(\{\textbf{x}^k \in \mathbb{R}^{1 \times M \times 2} | k = 1,2,3...J\}\), where \(J\) represents the number of frames in the input sequence. The goal is to optimize the prediction function (estimated 3D pose key points in three dimensions) \(f\) from the training data,
\begin{equation}
f(\textbf{x};W) = \arg \max_{y \in \mathcal{D}} F(\textbf{x},y;W)
\end{equation}
where \(W\) is the weight of the neural network and \(F\) is the optimizing function.
\label{subsec:formulation}

\subsection{Spatial Projection Guided Approach} \label{subsec:approach}

In the training process, SPGNet contains three components: an encoder, a projector and loss functions. First, the encoder processes the input 2D key points, aiming to transform the data into precise 3D coordinates that represent the estimated human pose. Then, the projector transforms the estimated 3D human pose into a re-projected 2D pose for computing the similarity with the original input. Finally, multiple loss functions guide the backpropagation of the neural network in order to learn helpful representations in the latent space.

\subsubsection{The Encoder} First, we encode our input data through an encoder and obtain the predicted 3D pose key points in three dimensions. Then, we decompose our output \(\hat{t}_{i}\) for a certain frame into \(\hat{t}_{i}^{k}\) and \(\hat{t}_{i}^{p}\), where \(\hat{t}_{i}^{k}\) is the predicted 3D pose key points coordinates in a certain 2D plane and \(\hat{t}_{i}^{p}\) is the spatial information to confirm the global position of pose. In this case, \(\hat{t}_{i}^{k} \in \mathbb{R}^{M \times 2}\), which is the \((x,y)\) coordinates of each joint of a 3D pose in three dimensions. Similarly, \(\hat{t}_{i}^{p} \in \mathbb{R}^{M}\) is just the \((z)\) coordinate as predicted by the position. We do same decomposition for label \(y_i\) and gain \(y_i^{k}\), \(y_i^{p}\) for further loss computation.

\begin{figure}[!ht]
\includegraphics[width=\textwidth]{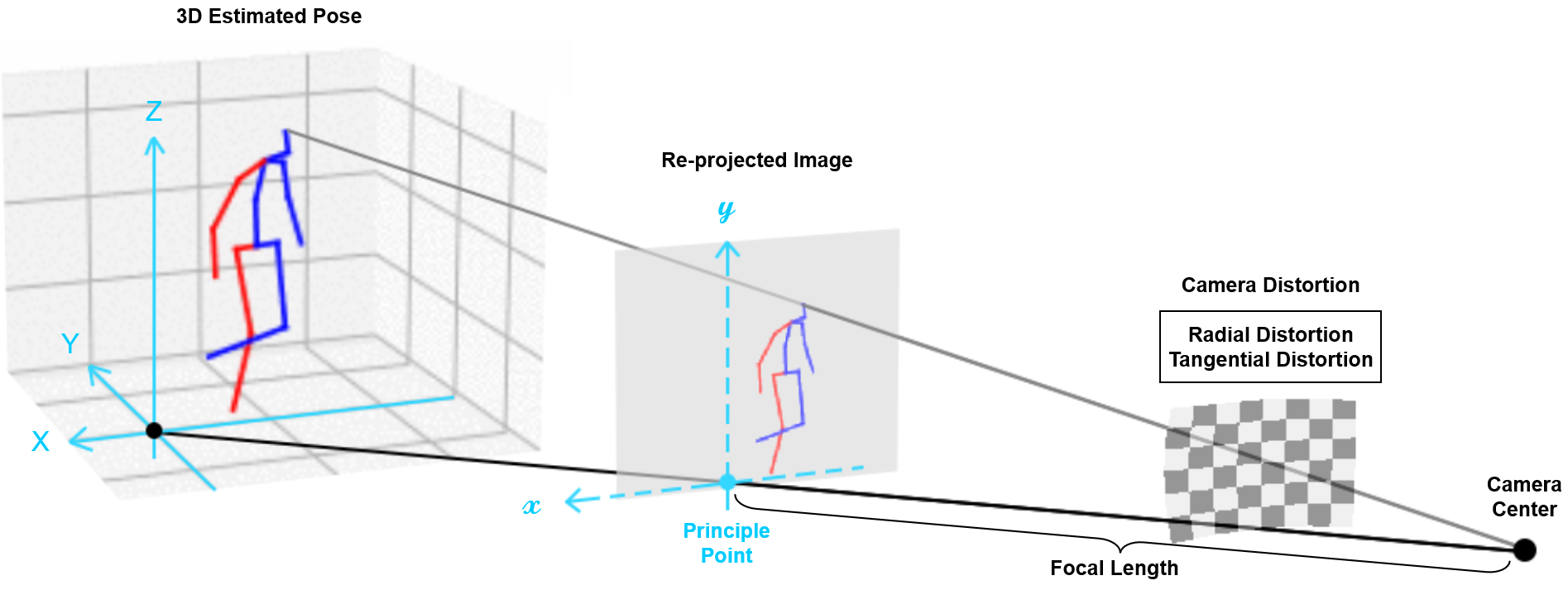}
\caption{Schematic diagram of the spatial dimension projection for SPGNet. The projector simulates the real camera to maintain consistency. The 3D estimated pose is projected to a 2D image with the same principle point, focal length, radial distortion and tangential distortion.} \label{projection}
\end{figure}

\subsubsection{The Projector} A projector is utilized to map the estimated 3D human pose into the re-projected 2D pose. In order to minimize the impact of lens distortion on projection, we chose the nonlinear projector. The schematic diagram is shown in Figure \ref{projection}. This schematic diagram shows the forward projection onto the image plane that maps \((x,y,z)\) coordinate into the \((x,y)\) plane. The center is the camera center, and the focal length is the distance between the camera center and the image plane along the principal axis perpendicular to the image plane. These two parameters represent the spatial geometry relationship of the 3D pose in low dimensions. Another essential aspect to consider is the camera distortion. It is a kind of optical aberration that causes a straight line in the scene to not remain straight in an image. There are two common camera distortions. First, the radial distortion causes the magnification of the image to decrease or increase with distance from the principal axis. Second, tangential distortion occurs because the lens assembly is not centered and parallel to the image plane. Fixing lens distortion during projection can increase projection accuracy. Consequently, the projector has several intrinsic camera parameters as input to ensure the projection is at a right angle, specified by the particular dataset. Here, we define focal length, center, radial distortion, and tangential distortion as \(f_c\), \(c_e\), \(d_r\), and \(d_t\), respectively. The pseudocode of nonlinear projection is summarized in Algorithm \ref{alg1}.

\begin{algorithm}
    \SetKwInOut{Input}{Input}
    \SetKwInOut{Output}{Output}

    \underline{Projector} $(\hat{t}_{i}^{k},\hat{t}_{i}^{p}, f_c,c_e,d_r,d_t)$\;
    \Input{Estimated 3D pose $\hat{t}_{i}^{k}$, estimated global position $\hat{t}_{i}^{p}$, intrinsic camera parameters $(f_c,c_e,d_r,d_t)$}
    \Output{Re-projected 2D pose $pose_{2d}$}
    
    $pose_{depth}$ = $\hat{t}_{i}^{k} / \hat{t}_{i}^{p}$ \;
    $pose_{cons}$ = Clamp ($pose_{depth}$) with $(min = -1, max = 1)$  \;
    $r$ = Sum ($pose_{cons}$) value between $(x,y)$ coordinates\;
    $concat$ = Concatenates the given sequence $(r, r^2, r^3)$ \;
    
    $R$ = Sum ($1 + (d_r \cdot concat)$) value between $(x,y,z)$ coordinates\;
    $T$ = Sum ($d_t \cdot pose_{cons}$) value between $(x,y)$ coordinates\;
    
    $pose_{trans}$ = $pose_{cons} \cdot (R + T) + d_t \cdot r $ \;
    $pose_{2d}$ = $f_c \cdot pose_{trans} + c_e$ \;
    \Return $pose_{2d}$
    
    \caption{Nonlinear projector mapping estimated 3D pose to re-projected 2D pose.}
    \label{alg1}
\end{algorithm}

\subsubsection{Loss functions} Usually, deep learning models predict the target without restraint. Thus, we propose the kinematic constraint loss as a penalty to maintain consistency. This is under the assumption that the length of human bones is constant from beginning to end. We construct the output 3D pose pairs \((\hat{\alpha}, \hat{\beta}\)), where \(\hat{\alpha}\) is the previous predicted frame \(\hat{t}\), and \(\hat{\beta}\) is current predicted frame. Then, we define kinematic constraint loss as follows:
\begin{equation}
\mathcal{L}_{kc}(\hat{\alpha};\hat{\beta}) =  \frac{1}{2M} \sum\limits_{i=1}^{M} \sum\limits_{j=1}^{P} abs( \Vert \hat{\alpha}_{i} - \hat{\alpha}_{j}  \Vert_2 - \Vert \hat{\beta}_{i} - \hat{\beta}_{j}  \Vert_2 )
\end{equation}
Where, \(\hat{\alpha}_j\) is the parent joint of the current joint \(\hat{\alpha}_i\). The 3D human pose can be regarded as a tree-like structure. Each joint has at least one connection with the other joint (parent), and some of them have multiple connections. Thus, for a particular joint, the number of connections is determined by the \(P\) denoted as parents. The dummy variable \(P\) depends on a specific dataset that consists of an indeterminate number of joints. The constant 0.5 is a multiplier for model stability because the repeated bone length would be calculated twice. Note that the first predicted frame does not have \(\hat{\alpha}\). Thus, we omit the computation of the loss function for the first predicted frame. This loss can be regarded as a penalty between consecutive frames and effectively maintaining the length of the bone, as shown in Section \ref{kcloss_d}.

The loss function for estimated 3D human pose is simply the mean per joint position error (MPJPE), which is the Protocol \(1\) used in many existing work: 
\begin{equation}
\label{eqn:MPJPE}
\mathcal{L}_{mpjpe}(\hat{t}_{i}^{k};y^{k}) =  \frac{1}{M}\sum\limits_{i=1}^{M} \Vert \hat{t}_{i}^{k} - y_i^{k}  \Vert_2
\end{equation}
The MPJPE loss will calculate the euclidean distance between all the joints of the predicted 3D pose and the ground truth. During the backpropagation, the loss gradient provides information for optimizing the degree of the key point's accuracy. Notably, the ground truth is in the camera space, transformed by using the intrinsic and extrinsic camera parameters. Therefore, for the global position \(y^{p}\), MPJPE cannot hold the depth information of the 3D pose. We use the weighted MPJPE loss function for estimated global position to retain the maximum spatial feature: 
\begin{equation}
\mathcal{L}_{wmpjpe}(\hat{t}_{i}^{p};y^{p}) =  \frac{1}{M}\sum\limits_{i=1}^{M} \frac{1}{y_i^{k}} \Vert \hat{t}_{i}^{p} - y_i^{k}  \Vert_2
\end{equation}
The inverse term \(1/y_i^{k}\) is the regularization term compared with MPJPE loss to force the predicted 3D pose centered around the trim area. This is assuming that the pose object cannot move very far away from the camera position. The model learns the 3D pose characteristics of centralization in terms of results. In addition, the weighted MPJPE loss significantly increases the accuracy of the projected 2D pose and reduces the error caused by the abnormal predicted global position. 

\subsection{Design of Encoder}
The architecture of the 2D-to-3D lifting neural network we designed is a temporal dilated convolutional model inspired by previous lifting approaches \cite{MartinezHRL17,temporal_convolutions_and_semi_supervised}. The neural network is fully implemented with the residual connections \cite{ResNet_2016} in order to transform the sequence of input \(\textbf{x}\) defined in \ref{subsec:formulation} through temporal convolutional layers. 
In detail, each residual blocks can be defined as  
\begin{equation}
\textbf{z} = D(\sigma (N(C(\textbf{x})))),
\end{equation}
where \textbf{z} is the extracted feature, \(C\) is the convolutional layer with 1024 input sizes except the input layer and 1024 output sizes, \(B\) is the batch normalization layer, \(\sigma\) is the ReLU activation layer, and finally, \(D\) is the dropout layer. Two residual blocks form a residual connection shown in Figure \ref{model}. The number of residual blocks depends on the number of input frames. For example, input \(\textbf{x}\) with \(J = 243\) needs 8 residual blocks for residual connections. Reminder, the input convolutional layer has \(2 \cdot M\) input size to adapt the 2D key points, where \(M\) is defined in Section \ref{subsec:formulation}. Finally, the output layer is a convolutional layer with output size \(3 \cdot M\), fitting with size of estimated 3D pose, defined as:
\begin{equation}
\hat{t}_{i} = C_{out}(\textbf{z}),
\end{equation}
where \(\hat{t}_{i}\) is the output of encoder for one frame defined in Section \ref{subsec:approach}.

\begin{table}[!htp]
\centering
\caption{Detailed results under Protocol 1 (MPJPE)}\label{tab1}
\setlength{\tabcolsep}{7.5pt} 
\renewcommand{\arraystretch}{1} 
\begin{threeparttable}
\begin{tabular}{|l|ccccccc|l|}
\hline 
Methods & Martinez& Sun & Yang & Lee & Pavllo & Cai & Xu&Our\\
& \cite{MartinezHRL17}&\cite{Sun_2017_ICCV}&\cite{Yang_2018_CVPR}&\cite{Lee_2018_ECCV}&\cite{temporal_convolutions_and_semi_supervised}&\cite{Cai_2019_ICCV}&\cite{Xu_2020_CVPR}&\\
\hline
Direct.&51.8&52.8&51.5&40.2&45.1&44.6&\textbf{37.4}&\underline{37.5}\\
Discuss.&56.2&54.8& 58.9&49.2&47.4& 47.4&\textbf{43.5}&\underline{44.7}\\
Eat& 58.1&54.2&50.4& 47.8&\underline{42.0}& 45.6&42.7&\textbf{41.8}\\
Greet&59.0&54.3& 57.0&52.6& 46.0& 48.8&\underline{42.7}&\textbf{42.1}\\
Phone &69.5& 61.8& 62.1&50.1&49.1& 50.8&\underline{46.6}&\textbf{45.5}\\
Photo &78.4&67.2&65.4&75.0&56.7& \underline{59.0}&59.7&\textbf{58.9}\\
Pose &55.2&53.1&49.8& 50.2&44.5&  47.2&\textbf{41.3}&\underline{42.0}\\
Purch. &58.1& 53.6&52.7&43.0& 44.4&\textbf{43.9}&\underline{45.1}&46.7\\
Sit&74.0&71.7&69.2&55.8&  57.2& 57.9&\textbf{52.7}&\underline{52.8}\\
SitD &94.6&86.7&85.2&73.9&66.1&61.9&\underline{60.2}&\textbf{59.4}\\
Smoke &62.3& 61.5&57.4& 54.1&47.5&49.7&\textbf{45.8}&\underline{46.7}\\
Wait &59.1&53.4&58.4& 55.6&44.8& 46.6&\underline{43.1}&\textbf{42.8}\\
WalkD &65.1&61.6&\textbf{43.6}&58.2&49.2& 51.3&47.7&\underline{46.6}\\
Walk &49.5&47.1& 60.1&43.3&32.6& 37.1&\textbf{33.7}&\underline{34.7}\\
WalkT &52.4&53.4& 47.7&43.3&34.0& 39.4&\underline{37.1}&\textbf{36.8}\\
\hline
Avg&62.9&59.1&58.6&52.8&47.1&48.8&\underline{45.6}&\textbf{45.3}\\
\hline
\end{tabular}
\begin{tablenotes}
        \footnotesize
        \item *The table reports the result with CPN 2D detection pose key points as input. The last line is the average of all 15 action results in millimeter. Best results in bold, second best underlined.
\end{tablenotes}
\end{threeparttable}
\end{table}

\section{Experiments}

\subsection{Dataset and Metrics}

We evaluate our method on the public dataset \textbf{Human3.6M}, widely used in other work \cite{h36m_pami}. The Human3.6M dataset is collected by a motion capture system. As one of the largest 3D human pose estimation datasets, 11 professional actors performed 15 scenarios consisting of 3.6 million video frames. There are four digital video cameras, one time-of-flight sensor, and ten motion cameras to capture the human pose. Our experiments follow the previous work \cite{temporal_convolutions_and_semi_supervised} to adopt a standard 17-joints skeleton and split the dataset into a training set (S1, S5, S6, S7, S8) and a testing set (S9, S11).

We take two widely used protocols to evaluate our model: \textbf{Protocol 1} is the mean per joint position error (MPJPE) defined in Equation \ref{eqn:MPJPE}. MPJPE calculates the Euclidean distance between the estimated positions of the joints and ground truth in millimeters. \textbf{ Protocol 2} is the Procrustes mean per joint position error (P-MPJPE), which calculates the error after aligning the estimated 3D pose to the ground truth in a rigid transformation, such as translation, rotation, and scale.

\subsection{Implementation Details}

We train our model with input in camera space for consistency of other work through quaternion transformation. Here, we set our temporal convolutional model with 243 frames to benefit from the consecutive video stream. We choose Adam \cite{adam} as an optimizer and train our model with 100 epochs. The learning rate starts from 0.001 and decays exponentially every epoch. We adopt fine-tuned 2D pose detection key points through the Cascaded Pyramid Network \cite{CPN} and ground truth 2D as our input. We apply the data augmentation method, pose flipping horizontally in the training set, with settings similar to \cite{temporal_convolutions_and_semi_supervised}.

\begin{table}[!htp]
\centering
\caption{Detailed results under Protocol 2 (P-MPJPE)}\label{tab2}
\setlength{\tabcolsep}{7.5pt} 
\renewcommand{\arraystretch}{1} 
\begin{threeparttable}
\begin{tabular}{|l|ccccccc|l|}
\hline 
Methods & Martinez& Sun & Yang & Lee & Pavllo & Liu & Xu&Our\\
& \cite{MartinezHRL17}&\cite{Sun_2017_ICCV}&\cite{Yang_2018_CVPR}&\cite{Lee_2018_ECCV}&\cite{temporal_convolutions_and_semi_supervised}&\cite{Liu_2020_CVPR}&\cite{Xu_2020_CVPR}&\\
\hline
Direct.&39.5&42.1&\textbf{26.9}&34.9&34.2&32.5&31.0&\underline{30.8}\\
Discuss.&43.2& 44.3& \textbf{30.9}&35.2& 36.8&35.3&36.8&\underline{34.6}\\
Eat& 46.4& 45.0&36.3&43.2&\textbf{33.9}&34.3&34.7&\underline{34.1}\\
Greet&47.0& 45.4& 39.9&42.6&37.5& 36.2&\textbf{34.4}&\underline{35.8}\\
Phone & 51.0&51.5& 43.9& 46.2& 37.1&37.8&\underline{36.2}&\textbf{35.3}\\
Photo &56.0& 53.0& 47.4& 55.0&43.2&\textbf{43.0}&43.9&\underline{43.2}\\
Pose & 41.4&43.2&\textbf{28.8}& 37.6& 34.4& 33.0&\underline{31.6}&31.9\\
Purch. &40.6&41.3&\textbf{29.4}& 38.8& 33.5&\underline{32.2}&33.5&32.5\\
Sit&56.5&59.3& \textbf{36.9}& 50.9&45.3&45.7&42.3&\underline{42.1}\\
SitD & 69.4&73.3&58.4& 67.3&  52.7& 51.8&\textbf{49.0}&\underline{49.9}\\
Smoke & 49.2& 51.0&41.5& 48.9&\underline{37.7}& 38.4&\textbf{37.1}&39.0\\
Wait & 45.0& 44.0&\textbf{30.5}&35.2&34.1&32.8&33.0&\underline{32.6}\\
WalkD & 49.5&48.0&\textbf{29.5}&50.7& 38.0&37.5&39.1&\underline{37.2}\\
Walk & 38.0&38.3&42.5&31.0& \textbf{25.8}&\textbf{25.8}&26.9&\underline{26.9}\\
WalkT & 43.1&44.8&32.2&34.6&\textbf{27.7}& \underline{28.9}&31.9&29.7\\
\hline
Avg&47.7&48.3&37.7&43.4&36.8&36.8&\underline{36.2}&\textbf{35.7}\\

\hline
\end{tabular}
\begin{tablenotes}
        \footnotesize
        \item *The table reports the result with CPN 2D detection pose key points as input. The last line is the average of all 15 action results in millimeter. Best in bold, second best underlined.
\end{tablenotes}
\end{threeparttable}
\end{table}

\subsection{Comparison with State-of-the-art Methods}

This section reports our model's performance on 15 actions belonging to S9 and S11. First, we use the CPN network as the 2D pose detector to obtain the 2D key points as our input data. We compare them using Protocol 1 and Protocol 2, shown in Table \ref{tab1} and \ref{tab2}. Our model has a lower average error than all other approaches under both protocols and does not rely on additional data as many other approaches. Under Protocol 1 (Table 1), our model slightly outperforms the previous best result \cite{Xu_2020_CVPR} by 0.3 mm on the average, corresponding to a 0.6\% error reduction. To be more specific, we got the best 7 out of 15 actions and 7 second best actions. This indicates that our model's architecture has a better learning ability to extract features in the latent space in order to keep the spatial and temporal information.
For Protocol 2, our model achieves the best results in terms of average P-MJPJE, with a 1.4\% error reduction. However, for individual actions, most actions predicted by our model only achieve the second-best results. Yang's \cite{Yang_2018_CVPR} work reports a significant improvement in the actions with complex spatial relationships, such as sitting or direction, achieving the seven best results. Compared to results in Table \ref{tab1}, their method uses the feature of GANs but makes it hard to detect the global position of the 3D human pose, leading the better performance in Protocol 1.

\begin{table}[!htp]
\centering
\caption{Detailed results based on ground truth of 2D human pose.}\label{tab3}
\setlength{\tabcolsep}{8pt} 
\renewcommand{\arraystretch}{1} 
\begin{threeparttable}
\begin{tabular}{|l|ccc |c|c|}
\hline 
 & & Protocol 1 &  &  & Protocol 2\\
\hline 
Methods &Hossain & Lee & Liu & Our & Our\\
&\cite{Hossain_2018} & \cite{Lee_2018_ECCV} & \cite{Liu_2020_CVPR} & &\\
\hline
Direct.&35.2&32.1&34.5&\textbf{29.5}&21.1\\
Discuss.&40.8& 36.6&37.1&\textbf{33.6}&23.9\\
Eat& 37.2& 34.3&33.6&\textbf{32.8}&24.5\\
Greet&37.4&37.8&34.2&\textbf{32.5}&24.6\\
Phone&43.2 &44.5&32.9&\textbf{32.5}&25.3\\
Photo&44.0&49.9& 37.1&\textbf{33.6}&29.9\\
Pose&38.9&40.9&39.6&\textbf{37.5}&21.1\\
Purch.&35.6 &36.2&35.8&\textbf{29.8}&23.7\\
Sit&42.3&44.1&40.7&\textbf{30.7}&30.4\\
SitD& 44.6& 45.6& 41.4&\textbf{38.1}&38.9\\
Smoke& 39.7&35.3& \textbf{33.0}&46.4&26.9\\
Wait& 39.7&35.9&\textbf{33.8}&35.9&23.1\\
WalkD&40.2&30.3&33.0&\textbf{31.8}&27.7\\
Walk& 32.8&37.6&26.6&\textbf{25.5}&18.8\\
WalkT&35.5& 35.5&\textbf{26.9}&28.5&19.6\\
\hline
Avg& 39.2&38.4&34.7&\textbf{33.4}&25.3\\

\hline
\end{tabular}
\begin{tablenotes}
        \footnotesize
        \item *Utilize the ground truth 2D human pose to predict the target. The table reports results of both Protocal 1 (left side) and Protocol 2 (right side). We label the best in bold.
\end{tablenotes}
\end{threeparttable}
\end{table}

To further study our method, we utilize the ground truth of 2D key points as our input to evaluate our model, with results shown in Table \ref{tab3} under Protocol 1 and Protocol 2. By using 2D ground truth, the models generally get better performance than Table \ref{tab2}. Compared to the previous best result \cite{Liu_2020_CVPR}, our model outperforms by 1.3 mm on the average. Our method highlights the reuse of the 2D key points for the 2D MPJPE loss, so the model depends more on the 2D key points than other models that only use those data at the input stage. This is evident from the higher error deduction comparing our model to previously implemented methods between Table \ref{tab2} and Table \ref{tab3}. More discussion about the improvement in performance in this case is presented in a later section. We also report our model's performance in Protocol 2, resulting in better performance of 25.3 mm on the average, outperforming the state-of-the-art.

\subsection{Ablation Study}

\subsubsection{Analysis of re-projected 2D pose loss} 
\begin{figure}[!htp]
\centering
\begin{subfigure}{.5\textwidth}
  \centering
  \includegraphics[width=\linewidth]{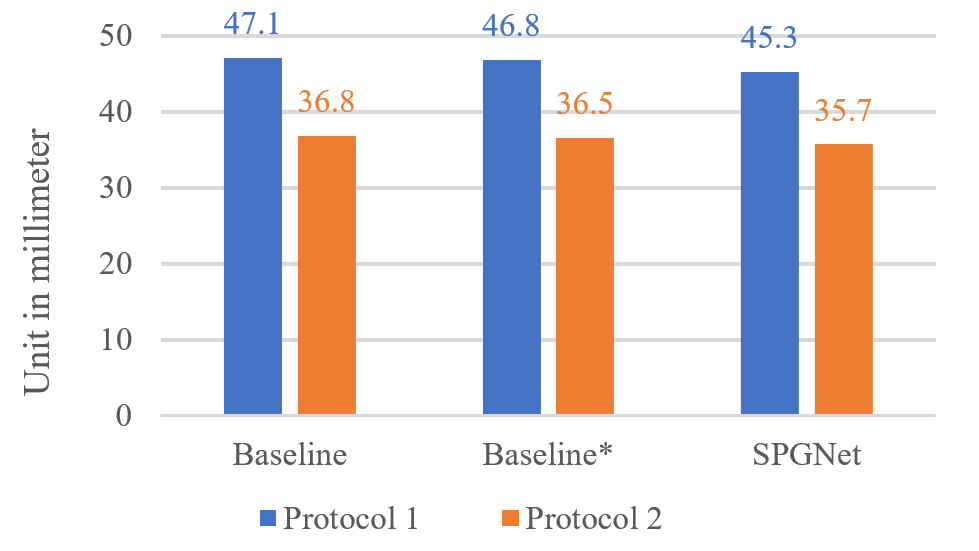}
  \caption{Comparisons of different settings}
  \label{fig:sub1}
\end{subfigure}%
\begin{subfigure}{.5\textwidth}
  \centering
  \includegraphics[width=\linewidth]{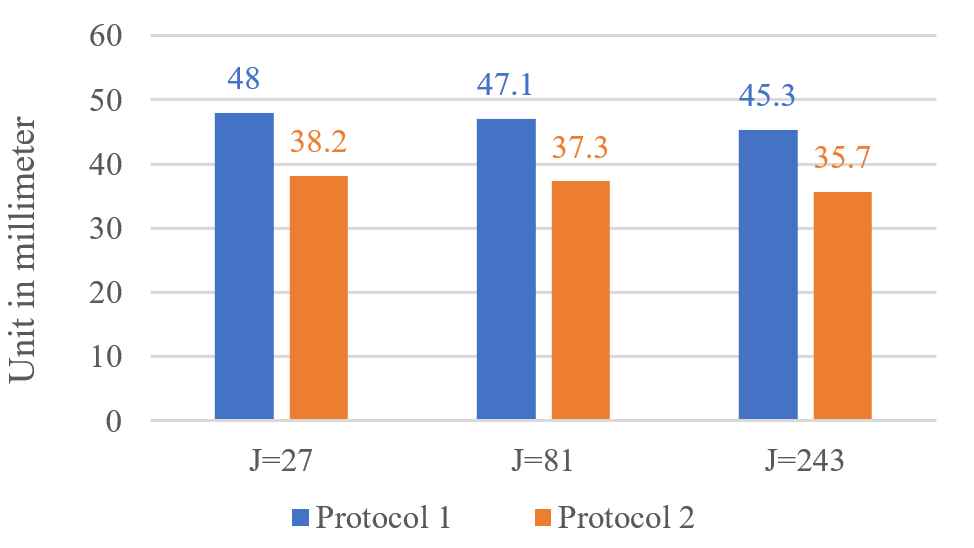}
  \caption{Comparisons of size of input frames}
  \label{fig:sub2}
\end{subfigure}
\caption{Clustered column charts (a) and (b) illustrate the comparisons of different sittings and input frames, respectively. The error shows a decreasing trend as the model is applied from the Baseline to our SPGNet or the size of input frames increases. Notably, in Figure \ref{fig:sub1}, Baseline is the model only learned from MPJPE loss, and Baseline* represents the Baseline adding kinematic constraints. In the chart \ref{fig:sub2}, \(J\) represents the number of frames in the input sequence.}
\label{fig:ablation}
\end{figure}

We further analyze the MPJPE loss between the re-projected 2D pose and the original input. We choose the encoder only consisting of MPJPE loss for the estimated 3D human pose as our encoder and add kinematic constraint as Baseline* to compare with SPGNet, as shown in Figure \ref{fig:sub1}. The histogram indicates that our model dramatically benefits from the MPJPE loss of the re-projected 2D pose, leading to the re-projected 2D pose being closer to the ground truth 2D human pose. Furthermore, the kinematic constraint slightly improves the model's accuracy, by adding a bone length constraint.

\subsubsection{Analysis of size of input frames} The size of the input frames has a significant impact on the pose estimation performance. For example, the chart in Figure \ref{fig:sub2} shows that the error in Protocol 1 decreases by 0.9 millimeters and 1.8 millimeters while the size of the input frames increases from 27 to 81 and 243, respectively. A similar error decrease trend is also reflected in Protocol 2. Consequently, we conclude that the larger size of sequential input provides more temporal information to allow our model to capture the movement of the 3D human pose between frames, similar to \cite{temporal_convolutions_and_semi_supervised}.

\subsection{Effect of kinematic constraints} \label{kcloss_d}

\begin{figure}[!htp]
\includegraphics[width=\textwidth]{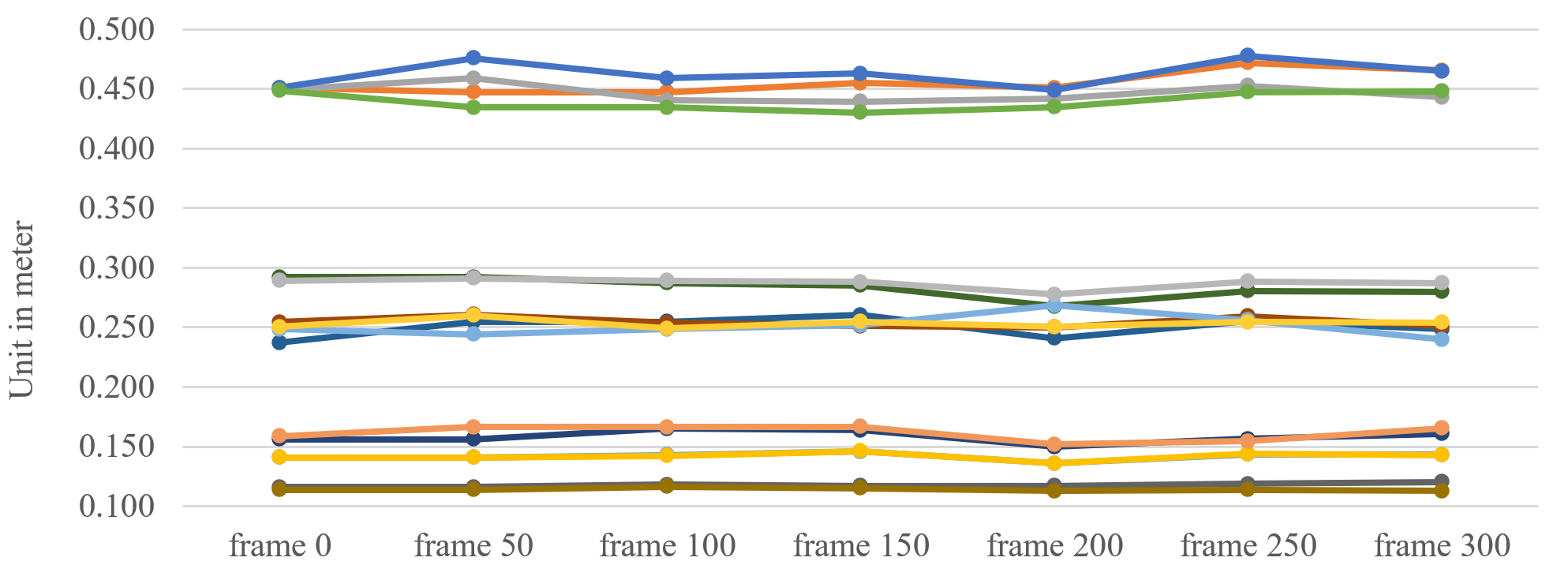}
\caption{Line chart of the measured bone lengths in seven different frames extracted from S11 walking scenarios. The different colors of lines represent the sixteen bones of a 3D human pose (tree-like structure).} \label{bones}
\end{figure}

In this section, we compare the sixteen bone lengths measured in different frames in one particular walking scenario, shown in Figure \ref{bones}. We can see that the difference in most small bones is almost negligible. Some large bones, such as leg bone or spine, have some relatively large variance compared to the small bone. However, the errors are all within 0.065 meters, which is acceptable and may be caused by movement of the frame. As a consequence, this line chart indicates that our kinematic constraint is effective.

\subsection{Shielding problem}
\begin{figure}[!htp]
\includegraphics[width=\textwidth]{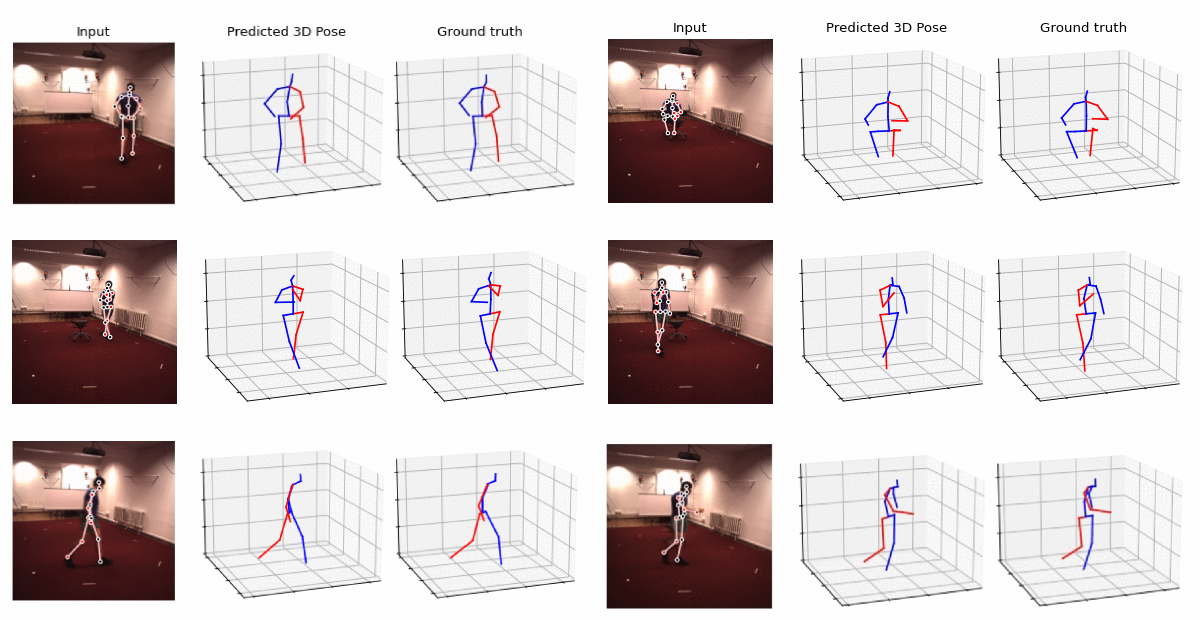}
\caption{Visualization of qualitative results on video clips.} \label{fig2}
\end{figure}

We further analyze the limitation of our method from visualization results, shown in Figure \ref{fig2}. We found that in the first row, which represents the action of sitting, the human limbs barely overlap in the camera's view. Thus, our results cannot be distinguished by the human eye if the consideration of global position is ignored. However, in the third row, the red arm of the human overlaps with the human body, which forms a shield masking the inner limbs. Consequently, the spatial relationship is not perfectly represented for both human pose in the third row, the variance of body tilt, and the difference of the arm's spatial positions, respectively.

\section{Conclusion}
We proposed SPGNet, a fully convolutional network based on supervised learning for human 3D pose estimation. To utilize our 2D-to-3d-lifting network, we used 2D key points in both input and re-projection stages and introduced kinematic constraints of human bone length and the corresponding loss function. Our model achieved more reliable estimates than state-of-the-art methods. SPGNet utilizes 2D labels in a more effective way, so the performance is expected to increase using image-to-2D methods with higher accuracies. Furthermore, besides the popular Human3.6M dataset, more datasets need to be tested for better analysis of our method.

%
%
%
\bibliographystyle{splncs04}
\bibliography{mybibliography}

\end{document}